\documentclass[letterpaper, 10 pt, conference]{ieeeconf}  
\IEEEoverridecommandlockouts                           
\usepackage{cite}
\usepackage{amsmath,amssymb,amsfonts}
\usepackage{algorithmic}
\usepackage{graphicx}
\usepackage{textcomp}
\usepackage{xcolor}
\usepackage{array}
\newcolumntype{P}[1]{>{\centering\arraybackslash}p{#1}}
\usepackage{makecell}
\usepackage[table,xcdraw]{xcolor}
\usepackage{multirow}
\usepackage{booktabs}
\usepackage{pifont}
\usepackage{colortbl}
\usepackage{subcaption}
\usepackage{bbding}
\usepackage{hyperref} 
\usepackage[capitalize]{cleveref}
\crefname{section}{Sec.}{Secs.}
\Crefname{section}{Section}{Sections}
\Crefname{table}{Table}{Tables}
\crefname{table}{Tab.}{Tabs.}
\definecolor{mygreen}{RGB}{114, 210, 126}
\newcolumntype{L}[1]{>{\raggedright\arraybackslash}p{#1}}
\usepackage{bm}
\renewcommand{\bm}[1]{\mathbf{#1}}

\title{\LARGE \bf
UltraHiT: A Hierarchical Transformer Architecture for Generalizable Internal Carotid Artery Robotic Ultrasonography
}
\author{Teng Wang$^{1,*}$, Haojun Jiang$^{1,*,\S}$, Yuxuan Wang$^{2,*}$, Zhenguo Sun$^{3}$, Xiangjie Yan$^{1}$, \\ Xiang Li$^{1}$ and Gao Huang$^{1,\dagger}$%
\thanks{This work is supported by Ministry of Industry and Information Technology of the People’s Republic of China (2024YFB4708200).}%
\thanks{$^{1}$Department of Automation, BNRist, Tsinghua University, Beijing, China. $^{2}$School of Computer Science and Technology, Xidian University, Xi'an, China. $^{3}$Beijing Academy of Artificial Intelligence, Beijing, China.}%
\thanks{$^{*}$These authors contributed equally to this work.}%
\thanks{$^{\S}$Haojun Jiang guided this work.}%
\thanks{$^{\dagger}$Corresponding author: Gao Huang. Email: gaohuang@tsinghua.edu.cn}%
}

\begin{document}

\maketitle
\thispagestyle{empty}
\pagestyle{empty}

\begin{abstract}
Carotid ultrasound is crucial for the assessment of cerebrovascular health, particularly the internal carotid artery (ICA).
While previous research has explored automating carotid ultrasound, none has tackled the challenging ICA.
This is primarily due to its deep location, tortuous course, and significant individual variations, which greatly increase scanning complexity.
To address this, we propose a \underline{Hi}erarchical \underline{T}ransformer-based decision architecture, namely UltraHiT, that integrates high-level variation assessment with low-level action decision.
Our motivation stems from conceptualizing individual vascular structures as morphological variations derived from a standard vascular model.
The high-level module identifies variation and switches between two low-level modules: an adaptive corrector for variations, or a standard executor for normal cases.
Specifically, both the high-level module and the adaptive corrector are implemented as causal transformers that generate predictions based on the historical scanning sequence.
To ensure generalizability, we collected the first large-scale ICA scanning dataset comprising 164 trajectories and 72K samples from 28 subjects of both genders.
Based on the above innovations, our approach achieves a 95\% success rate in locating the ICA on unseen individuals, outperforming baselines and demonstrating its effectiveness.
Our code will be released after acceptance.
\end{abstract}

\section{introduction}

\begin{figure}[t!]
\centering
\includegraphics[width=1\columnwidth]{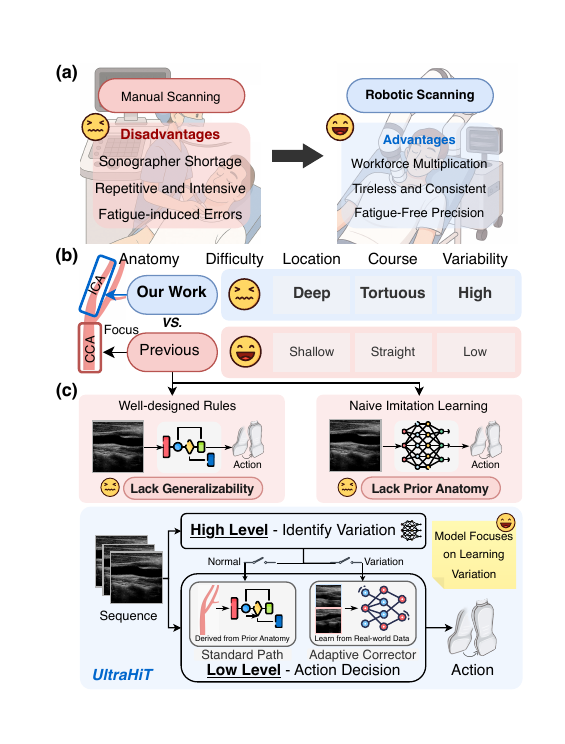}
\caption{
\textbf{Overview}.
(a) Robotic scanning vs. manual scanning.
(b) Characteristics of ICA vs. CCA.
(c) Our hierarchical transformer architecture vs. previous works.
}
\label{fig:overview}
\vspace{-15pt}
\end{figure}

Ultrasonography is the preferred method for assessing carotid artery health due to its real-time imaging, radiation-free, and cost-effectiveness features.
The procedure requires operators to precisely adjust the probe’s angle and position to obtain optimal imaging planes, demanding not only anatomical knowledge and interpretation skills for ultrasound images but also operational proficiency.
Training an experienced sonographer typically takes several years, resulting in a shortage of professionals in underdeveloped regions.
Moreover, ultrasound scanning is a highly repetitive and intensive task, where prolonged operation may lead to fatigue and compromised image quality \cref{fig:overview}(a).
These factors drive the development of autonomous ultrasound robots.

Carotid ultrasound robots are expected, with perception and decision-making capabilities, to autonomously adjust the probe based on real-time imaging, thereby obtaining clear transverse and longitudinal views of the vessels.
Current researches primarily focus on the automated scanning of the common carotid artery (CCA).
Specifically, researchers~\cite{huang2024robot,wang2024autonomous,yan2024unified,huang2021towards,huang2023motion,goel2022autonomous,duan2022ultrasound,yan2023multi} attempted to embed the knowledge and skills which required for carotid ultrasound into sophisticated rules, known as rule-based methods.
By analyzing image changes resulting from specific probe adjustment actions, they establish a set of mapping rules from image features to motion decisions.
Nevertheless, their reliance on an identical vascular model and neglect of anatomical variability limit their generalization ability.
The ICA's substantial variability makes creating an exhaustive rule set nearly impossible.

\begin{figure*}[htp!]
\centering
\includegraphics[width=2\columnwidth]{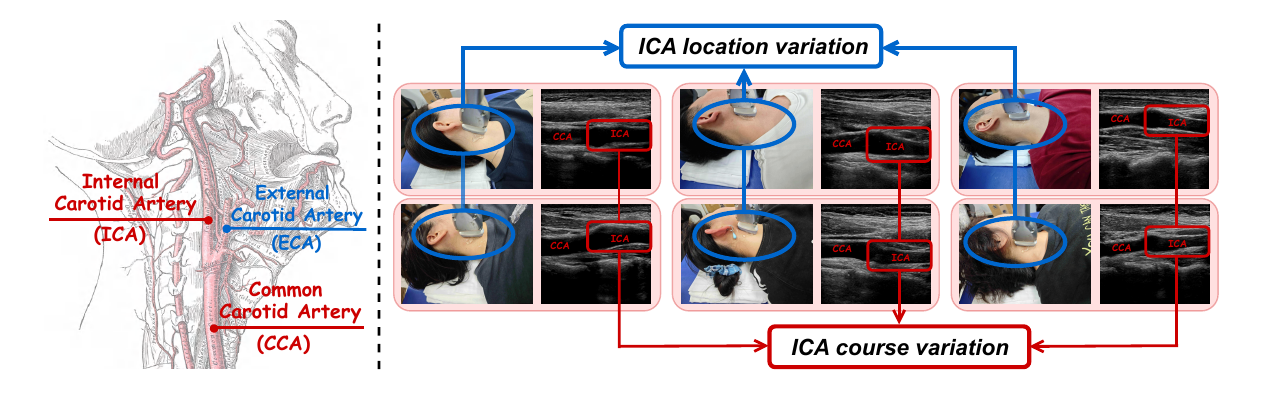}
\caption{
\textbf{Internal carotid artery's anatomy and variability}. 
The left panel shows the standard anatomy of the ICA~\cite{gray1878anatomy}, while the right panel demonstrates the significant variability in the position and course of the ICA within the population.
}
\label{fig:ica_illustration}
\vspace{-10pt}
\end{figure*}

Inspired by recent advances in deep learning, another line of research~\cite{jiang2025towards,deng2021learning,droste2020automatic,bi2025gaze,jiang2024cardiac,bi2022vesnet,li2021autonomous,li2023rl,jiang2024sequence} attempts to adopt learning-based methods for better generalization capability.
Learning-based methods include reinforcement and imitation learning.
Reinforcement learning method with either virtual vessel models~\cite{bi2022vesnet} or 3D CT data of organs~\cite{bi2024autonomous,li2021autonomous,li2023rl} is used to simulate the environment.
However, while the former suffers from poor generalization due to limited anatomical variations, the latter relies on costly and hard-to-scale 3D CT data.
Another category of methods~\cite{jiang2025towards,deng2021learning,droste2020automatic,bi2025gaze} uses imitation learning, which collects expert operation data from real-world individuals and trains models to imitate expert decisions.
Recently, work~\cite{jiang2025towards} demonstrated that the imitation learning strategy exhibits good generalization potential.

Despite these developments, existing automation efforts~\cite{jiang2025towards,huang2024robot,yan2024unified,huang2021towards,huang2023motion,bi2022vesnet,bi2025gaze} have primarily focused on the CCA, overlooking the clinically more significant and challenging ICA (\cref{fig:overview}b).
To enhance carotid robots, this paper targets the ICA—a primary brain blood pathway extending from the CCA (\cref{fig:ica_illustration} left).
Therefore, the ICA represents a critical window for evaluating cerebral blood supply, and obtaining its longitudinal view is particularly essential in carotid artery examinations.
However, ultrasonographic scanning of the ICA is more challenging than that of the CCA, primarily due to significant individual variations in: (1) vascular course, and (2) positional anatomy.
\cref{fig:ica_illustration} illustrates ICA conditions in six subjects: ultrasound images reveal differences in its course, while camera images show variations in its location.
This requires the ultrasound robot to possess strong adaptive capabilities, enabling it to recognize individual anatomical differences in real-time and make adjustment decisions, which presents a significant challenge.

To address this, as shown in \cref{fig:overview}c, we propose a \textbf{Hi}erarchical \textbf{T}ransformer-based decision architecture named \textbf{UltraHiT}, which integrates high-level variation identification with low-level action decision.
Our core idea is to treat individual vascular structures as variations derived from a standard vascular template.
We first construct a high-level state assessment module to identify whether vascular variations are present (\textit{i.e.}, whether adaptive adjustment is needed).
Furthermore, based on the high-level semantic signal, the system activates one of two low-level decision executors:
\textbf{(1)} A standardized path executor, based on anatomical knowledge from the standard ICA vascular model.
This standard vascular model inherently encapsulates common anatomical knowledge across populations, embodying a knowledge-enhanced approach that handles basic and normal scenarios effectively.
Equipped with this prior knowledge, the learning-based component is freed from learning basic structural information and can be tailored to focus on capturing vascular variations, thereby enhancing its generalization capability.
\textbf{(2)} An adaptive corrector, trained through imitation learning on large-scale data to capture potential individual variations.

Specifically, the high-level module and the adaptive corrector are both implemented as causal transformers. 
They are trained independently, each with its own distinct supervisory signal tailored to their specific tasks.
Furthermore, we posit that incorporating historical scan data provides richer structural information about the subject, thereby facilitating more informed decision-making. 
Consequently, the model input is extended to include a sequence of historical scans rather than a single image~\cite{jiang2025towards,huang2024robot,huang2021towards,bi2022vesnet}.
Finally, to train the model and ensure its generalizability, we collected the first large-scale ICA scanning dataset comprising 164 trajectories and 72K samples from 28 subjects of both genders.

To validate the effectiveness of our model, we conducted real-world experiments by testing the scanning success rate on unseen subjects of both gender.  
We compared our approach with rule-based and imitation learning methods, and the results demonstrate the efficacy of our proposed model.  
Furthermore, failure case analysis shows that our model does not exhibit issues such as complete deviation from the target, mis-targeting, or timeout before task completion, which are commonly observed in baseline models.  
Additionally, our method demonstrates stronger robustness under challenging conditions, including poor initial states and subject movement.  
In summary, the contributions are three fold:     
\begin{itemize} 
    \item We are the first to achieve autonomous scanning of the ICA longitudinal section as required in clinical practice, extending the capabilities of carotid robots.
    \item We propose a novel hierarchical transformer-based decision architecture named UltraHiT, which effectively handles anatomical variations in the ICA.  
    \item We have collected the first expert demonstration dataset for ICA scanning, comprising 164 trajectories and 72K samples from 28 subjects of both genders.
\end{itemize}
    
\section{Related Works}
\subsection{Rule-based Ultrasound System}
One category of researchers employs rule-based methods~\cite{huang2024robot,wang2024autonomous,yan2024unified,huang2021towards,huang2023motion,goel2022autonomous,duan2022ultrasound,yan2023multi}.
For instance, some works~\cite{huang2024robot,huang2021towards,wang2024autonomous} designed visual servoing methods based on the relationship between vascular images and the angle/position of the probe.
Yan et al.~\cite{yan2023multi,yan2024unified} utilized external cameras to identify neck structures and planned scanning trajectories based on the recognition results.
Such approaches are generally built upon a standard vascular model that captures common structural features across the population, thereby achieving a certain level of adaptability.
However, due to significant variability in individual vasculature, the generalization capability of rule-based approaches is fundamentally limited.

\subsection{Learning-based Ultrasound System}
Another category of work is the learning-based approach that has emerged in recent years~\cite{jiang2025towards,deng2021learning,droste2020automatic,bi2025gaze,jiang2024cardiac,bi2022vesnet,li2021autonomous,li2023rl,jiang2024sequence}.
For instance, Bi et al.~\cite{bi2022vesnet} created a virtual vascular simulation environment for training reinforcement learning models, but the vessel models are idealized and lack realistic anatomical variations.
In contrast, others~\cite{bi2024autonomous, li2021autonomous,li2023rl} used individual 3D CT scan data to construct the simulation environment—such data encompass real-world variations, yet are costly to acquire and difficult to scale.
Although reinforcement learning holds potential for learning optimal scanning policies, significant challenges in simulating realistic environments hinder its further development.
Another line of research follows the imitation-learning approach.
For instance, Jiang et al.~\cite{jiang2025towards} collected a large-scale expert scanning dataset to enable imitation learning and demonstrated its effectiveness in scanning individuals with carotid plaques, highlighting the potential of the method;
Deng et al.~\cite{deng2021learning} and Droste et al.~\cite{droste2020automatic} also validated the feasibility of this technology in abdominal and fetal imaging, respectively.
While such methods rely entirely on learning common structures and individual variations from data, the common anatomical features have already been well-established as medical knowledge. Relearning them from data results in lower data utilization efficiency.
The strength of learning-based methods lies in their ability to effectively model variations. Therefore, the focus of learning should be placed on capturing individual-specific variability.
Furthermore, existing methods have primarily focused on the automatic scanning of the common carotid artery (CCA). Exploration of the internal carotid artery (ICA), which is more challenging and clinically significant, remains insufficient.
\begin{figure*}[t!]
\centering
\includegraphics[width=2\columnwidth]{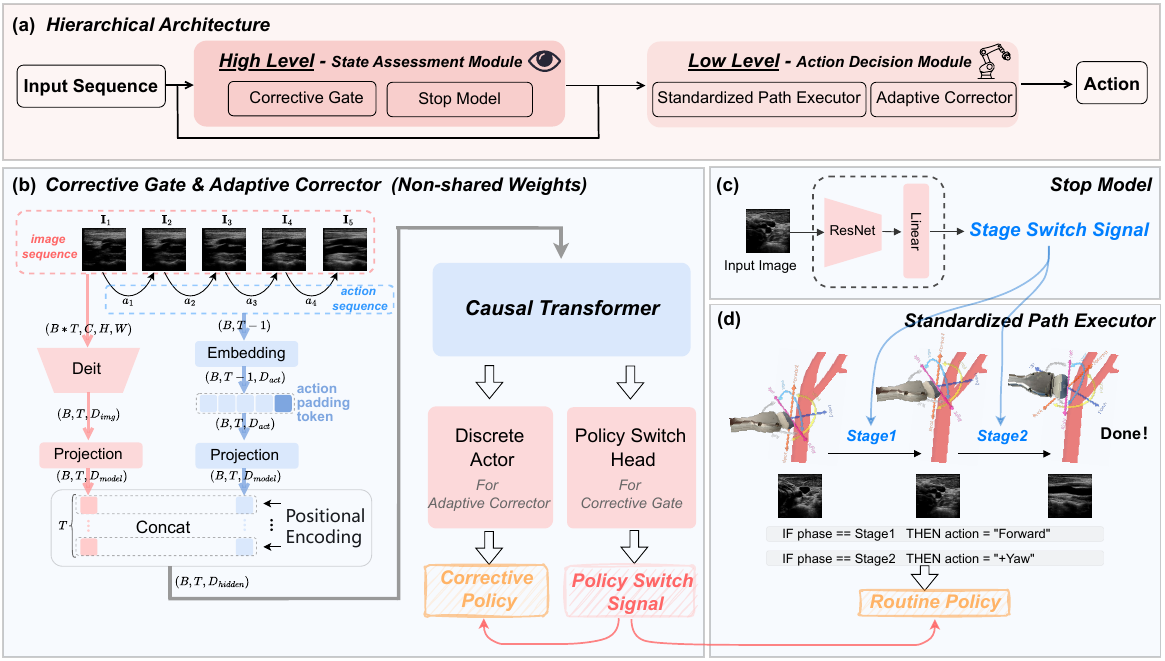}
\caption{
\textbf{Hierarchical transformer architecture}. 
(a) Overview of hierarchical architecture. The high-level module makes semantic decisions, while the low-level module executes physical actions in the real world.
(b) The corrective gate and adaptive corrector, process image-action sequences through a causal transformer.
(c) The stop model architecture.
(d) The standardized path executor, a knowledge-based policy designed using anatomical prior knowledge.
}
\label{fig:method}
\vspace{-10pt}
\end{figure*}

\section{method}

To address significant individual variations in ICA scanning, we propose a \textbf{Hi}erarchical \textbf{T}ransformer-based decision architecture, named \textbf{UltraHiT}. Our approach conceptualizes the scanning process as a sequence of high-level state assessments and low-level action decisions. This hierarchical design allows the system to leverage a knowledge-based standard scanning path for common anatomies while employing a data-driven adaptive model to handle complex variations. The overall architecture is shown in \cref{fig:method}.

\subsection{High-Level State Assessment Module}
\textbf{Corrective Gate}.
The Corrective Gate is designed to assess the current scanning trajectory and determine whether vascular variations are present, indicating the need for adaptive correction.
It is implemented as a Causal Transformer~\cite{chen2021decision} which takes a sequence of historical data $H_t$ (including past images and actions) as input to generate a binary policy switch signal $g_t \in \{0, 1\}$:
\begin{equation} 
g_t = \mathcal{H}_{\mathrm{gate}}(\mathcal{T}_{\mathrm{gate}}(H_t))
\end{equation} 
where $\mathcal{T}_{\mathrm{gate}}$ represents the Causal Transformer backbone and $\mathcal{H}_{\mathrm{gate}}$ is the dedicated Policy Switch Head. $g_t=1$ activates the Adaptive Corrector, while $g_t=0$ activates the Standardized Path Executor. The detailed architecture of the Causal Transformer $\mathcal{T}_{\mathrm{gate}}$ will be elaborated in \cref{causal}.

\textbf{Stop Model}.
The Stop Model is responsible for identifying the completion of scanning stages. As shown in \cref{fig:method}(c), it processes only the current ultrasound image $\mathbf{I}_t \in \mathbb{R}^{C \times H \times W}$ to generate a stage switch signal. We utilize a ResNet architecture, $\Phi_{\mathrm{ResNet}}$, pre-trained on ImageNet, as the feature extractor, followed by a linear layer to make the final decision. The process can be formulated as:
\begin{equation} 
s = \text{Softmax}(\mathbf{W}_{\mathrm{stop}} \cdot \Phi_{\mathrm{ResNet}}(\mathbf{I}_t) + \mathbf{b}_{\mathrm{stop}})
\end{equation} 
where $\mathbf{W}_{\mathrm{stop}}$ and $\mathbf{b}_{\mathrm{stop}}$ are the weight and bias of the final linear layer, and $s$ is the probability distribution over stage completion signals.

\subsection{Low-Level Action Decision Module}
\textbf{Standardized Path Executor}.
As illustrated in \cref{fig:method}d, the Standardized Path Executor is a deterministic policy based on anatomical prior knowledge of carotid arteries.
The ICA runs posterior-laterally while the ECA runs anterior-medially (\cref{fig:ica_illustration}(left)).
At the CCA bifurcation, the ICA is more lateral, making it the first to be seen when the probe rotates clockwise around the z-axis.
Based on this anatomical configuration, we define the action $a_t$ in each state $s_t$ as:
\begin{equation} 
a_t = 
\begin{cases}
\text{Forward}, & \text{if } s_t \in \text{Stage1}, \\
\text{+Yaw}, & \text{if } s_t \in \text{Stage2}. \\
\end{cases}
\end{equation} 
where ``Forward'' advances along the local x-axis
``+Yaw'' rotates clockwise around the z-axis (\cref{fig:method}d).
This routine policy provides an effective and reliable scanning strategy based on anatomical knowledge, 
thereby alleviating the learning burden on data-driven corrective model.

\textbf{Adaptive Corrector}.
The Adaptive Corrector is the core component for handling anatomical variability. It is a data-driven policy learned from expert demonstrations. Its goal is to generate corrective actions that bring a deviating scanning path back to the optimal trajectory. It utilizes a Causal Transformer architecture~\cite{chen2021decision} to map the state history $H_t$ to a corrective policy $\pi_{t}^{\mathrm{corr}}$:
\begin{equation} 
\pi_{t}^{\mathrm{corr}} = \mathcal{H}_{\mathrm{corr}}(\mathcal{T}_{\mathrm{corr}}(H_t))
\end{equation} 
where $\mathcal{T}_{\mathrm{corr}}$ is the Causal Transformer and $\mathcal{H}_{\mathrm{corr}}$ is the actor head that outputs a probability distribution over the action space. The action with the highest probability is then selected for execution. 
The action space consists of 12 discrete actions, corresponding to 12 different movement directions. 
These movements include translations and rotations along the x, y, and z axes of the probe's coordinate system.

\subsection{Causal Transformer for Decision Making} 
\label{causal}
Both the Corrective Gate and Adaptive Corrector are based on a Causal Transformer backbone (\cref{fig:method}(b)), which models temporal dependencies across ultrasound images and past actions to inform future decisions.

\textbf{Input Representation and Embedding}. 
The input consists of a sequence of the last $T$ ultrasound images $\mathcal{I} = \{\mathbf{I}_1, \dots, \mathbf{I}_T\}$ and $T-1$ past actions $\mathcal{A} = \{a_1, \dots, a_{T-1}\}$.

Each image $\mathbf{I}_t$ is encoded via a Vision Transformer (DeiT, $\Phi_{\mathrm{DeiT}}$) into a feature vector, then projected into $\mathbb{R}^{D_{\mathrm{model}}}$:
\begin{equation} 
\mathbf{f}_t^{\mathrm{img}} = \Phi_{\mathrm{DeiT}}(\mathbf{I}_t) \in \mathbb{R}^{D_{\mathrm{img}}} 
\label{eq:img_feat}
\end{equation} 
\begin{equation} 
\mathbf{x}_t^{\mathrm{img}} = \mathbf{f}_t^{\mathrm{img}} \mathbf{W}_{\mathrm{img\_proj}} \in \mathbb{R}^{D_{\mathrm{model}}} 
\label{eq:img_proj}
\end{equation} 
where $\mathbf{W}_{\mathrm{img\_proj}} \in \mathbb{R}^{D_{\mathrm{img}} \times D_{\mathrm{model}}}$ is a projection matrix. 

Past actions are embedded using a learnable matrix $\mathbf{E}_{\mathrm{act}}$. 
An action padding token $\mathbf{a}_{\mathrm{pad}}$ is appended for length alignment, and the sequence is projected via an MLP $\Phi_{\mathrm{act\_proj}}$:
\begin{equation} 
\mathbf{e}_t = 
\begin{cases} 
\mathbf{E}_{\mathrm{act}}[a_t] & \text{if } 1 \leq t \leq T-1 \\ 
\mathbf{a}_{\mathrm{pad}} & \text{if } t = T 
\end{cases} 
\label{eq:action_embed}
\end{equation} 
\begin{equation} 
\mathbf{x}_t^{\mathrm{act}} = \Phi_{\mathrm{act\_proj}}(\mathbf{e}_t) \in \mathbb{R}^{D_{\mathrm{model}}} 
\label{eq:action_proj}
\end{equation} 

\textbf{Sequence Modeling with Causal Transformer}. 
At each step $t$, image and action features are concatenated into a token $\mathbf{z}_t \in \mathbb{R}^{D_{\mathrm{hidden}}}$ ($D_{\mathrm{hidden}} = 2 \cdot D_{\mathrm{model}}$). Learnable positional embeddings $\mathbf{P}$ are added:
\begin{equation} 
\mathbf{H}_0 = [\mathbf{z}_1, \mathbf{z}_2, \dots, \mathbf{z}_T]^T + \mathbf{P}_{1:T} \in \mathbb{R}^{T \times D_{\mathrm{hidden}}} 
\label{eq:h0}
\end{equation} 

The sequence $\mathbf{H}_0$ is then processed by $L$ Causal Transformer blocks. Each block consists of a Causal Multi-Head Self-Attention (MHSA) layer and an MLP with pre-normalization. For the $l$-th block: 
\begin{equation} 
\mathbf{H}'_l = \text{LayerNorm}(\mathbf{H}_{l-1}) 
\label{eq:ln1}
\end{equation} 
\begin{equation} 
\mathbf{H}''_l = \text{MHSA}(\mathbf{H}'_l) + \mathbf{H}_{l-1} 
\label{eq:mhsa_res}
\end{equation} 
\begin{equation} 
\mathbf{H}_{l} = \text{MLP}(\text{LayerNorm}(\mathbf{H}''_l)) + \mathbf{H}''_l 
\label{eq:mlp_res}
\end{equation} 

The causal MHSA uses $h$ heads. For each head $i$, queries $\mathbf{Q}_i$, keys $\mathbf{K}_i$, and values $\mathbf{V}_i$ are derived via linear projections. 
The attention scores are computed using scaled dot-product attention with a causal mask $\mathbf{M}$: 
\begin{equation} 
\text{Attention}(\mathbf{Q}_i, \mathbf{K}_i, \mathbf{V}_i) = 
\text{softmax}\left(\frac{\mathbf{Q}_i \mathbf{K}_i^T}{\sqrt{d_k}} + \mathbf{\tilde{M}}\right) \mathbf{V}_i 
\label{eq:attention}
\end{equation} 
The causal mask $\mathbf{\tilde{M}}$ is a lower-triangular matrix where $\tilde{M}_{ij} = 0$ if $j \le i$ and $\tilde{M}_{ij} = -\infty$ otherwise, ensuring that the prediction at time step $t$ only depends on past inputs. 

The outputs of all heads are concatenated and projected back to the hidden dimension: 
\begin{equation} 
\text{MHSA}(\mathbf{H}'_l) = \text{Concat}(\text{head}_1, \dots, \text{head}_h) \mathbf{W}^O 
\label{eq:mhsa_out}
\end{equation} 
where $\text{head}_i = \text{Attention}(\mathbf{Q}_i, \mathbf{K}_i, \mathbf{V}_i)$ and $\mathbf{W}^O \in \mathbb{R}^{D_{\mathrm{hidden}} \times D_{\mathrm{hidden}}}$ is the output projection matrix.

\textbf{Output Heads}. 
After the final transformer block, we apply layer normalization and use only the output token at the last time step $\mathbf{h}_{\mathrm{final}}$ for decision-making. This vector encapsulates the information from the entire history. 
This final hidden state is then fed into task-specific heads: 
\begin{itemize} 
\item For the Adaptive Corrector, a discrete actor head outputs the logits for the corrective action policy: 
\begin{equation} 
\pi_{\mathrm{corr}} = \text{Softmax}(\mathbf{h}_{\mathrm{final}} \mathbf{W}_{\mathrm{actor}} + \mathbf{b}_{\mathrm{actor}}) 
\label{eq:actor}
\end{equation} 
\item For the Corrective Gate, a policy switch head outputs the logits for the gate signal: 
\begin{equation} 
g_{\mathrm{logits}} = \mathbf{h}_{\mathrm{final}} \mathbf{W}_{\mathrm{gate}} + \mathbf{b}_{\mathrm{gate}} 
\label{eq:gate}
\end{equation} 
\end{itemize} 

The Causal Transformer backbone weights are not shared between the Corrective Gate and Adaptive Corrector, enabling task specialization. Both models are trained independently using separate supervisory signals.

\subsection{Robot Control Algorithm}

We employ Cartesian impedance control \cite{albu2002cartesian}, following the same strategy of \cite{jiang2025towards}, to balance accuracy and compliance during scanning.
The joint dynamics of the Franka robotic arm can be described as:
\begin{equation}
    \mathbf{M}(\mathbf{q})\ddot{\mathbf{q}} + \mathbf{C}(\bm q,\dot{\bm q})\dot{\bm q} + \bm g(\bm q) = \bm \tau + \bm \tau_{\mathrm{ext}},
\end{equation}
where $\mathbf{M}$, $\mathbf{C}$, and $\mathbf{g}$ denote the mass matrix, Coriolis/centrifugal forces, and gravity vector, respectively. 
$\bm \tau$ is the commanded torque, and $\bm \tau_{\mathrm{ext}}$ represents external torques arising from probe–neck contact.

The impedance controller generates torques:
\begin{align}
\bm{\tau} &= \bm{J}^T(-\mathbf{K}\tilde{\bm{x}} - \mathbf{D}\dot{\bm{x}}) \nonumber \\
&\quad + (\mathbf{I} - \mathbf{J}^T \mathbf{J}^{+T})(-\mathbf{K}_n \dot{\bm{q}} - \mathbf{D}_n \tilde{\bm{q}}) 
+ \mathbf{C}\dot{\bm{q}} + \bm{g},
\end{align}
where $\tilde{\bm{x}}=\bm{x}-\bm{x}_d$ is the Cartesian pose error, $\bm{J}$ is the Jacobian, 
$\mathbf{K}$ and $\mathbf{D}$ are task-space stiffness and damping matrices, 
$\mathbf{K}_n$, $\mathbf{D}_n$ provide null-space regulation, and the superscript $+$ indicates the pseudo-inverse.

Substituting into the dynamics yields the closed-loop form:
\begin{equation}
\bm M \ddot{\bm q} + \bm J^T (\bm K \tilde{\bm x} + \bm D \dot{\bm x})
+ (\bm I - \bm J^T \bm J^{+T})(\bm K_n \dot{\bm q} + \bm D_n \tilde{\bm q})
= \bm \tau_{\mathrm{ext}}.
\end{equation}

Mapping to Cartesian space:
\begin{equation}
\bm J \bm M \ddot{\bm q} + \bm J \bm J^T (\bm K \tilde{\bm x} + \bm D \dot{\bm x}) = \bm F_{\mathrm{ext}},
\end{equation}
where $\bm F_{\mathrm{ext}} = \bm J^{+T} \bm \tau_{\mathrm{ext}}$ is the contact force.
In quasi-static contact ($\ddot{\bm q}\!\approx\!0, \dot{\bm x}\!\approx\!0$), this simplifies to:
\begin{equation}
\bm F_{\mathrm{ext}} = \bm K \tilde{\bm x},
\end{equation}
indicating that large pose errors may generate excessive contact forces, posing potential safety risks.

To improve safety, we adopt an error-dependent stiffness:
\begin{equation}
k =
\begin{cases}
k_{\text{normal}}, & f_{\mathrm{ext}} < \bar{f}_{\mathrm{ext}}, \\
\bar{\bm f}_{\mathrm{ext}}/\tilde{\bm x}, & f_{\mathrm{ext}} \ge \bar{f}_{\mathrm{ext}},
\end{cases}
\end{equation}
where $\bar{\bm f}_{\mathrm{ext}}$ is a predefined safe force threshold vector. 
This adjustment reduces stiffness when excessive forces are detected, limiting contact pressure on the patient. 
\section{Experiments}

\begin{figure}[t!]
\centering
\includegraphics[width=1\columnwidth]{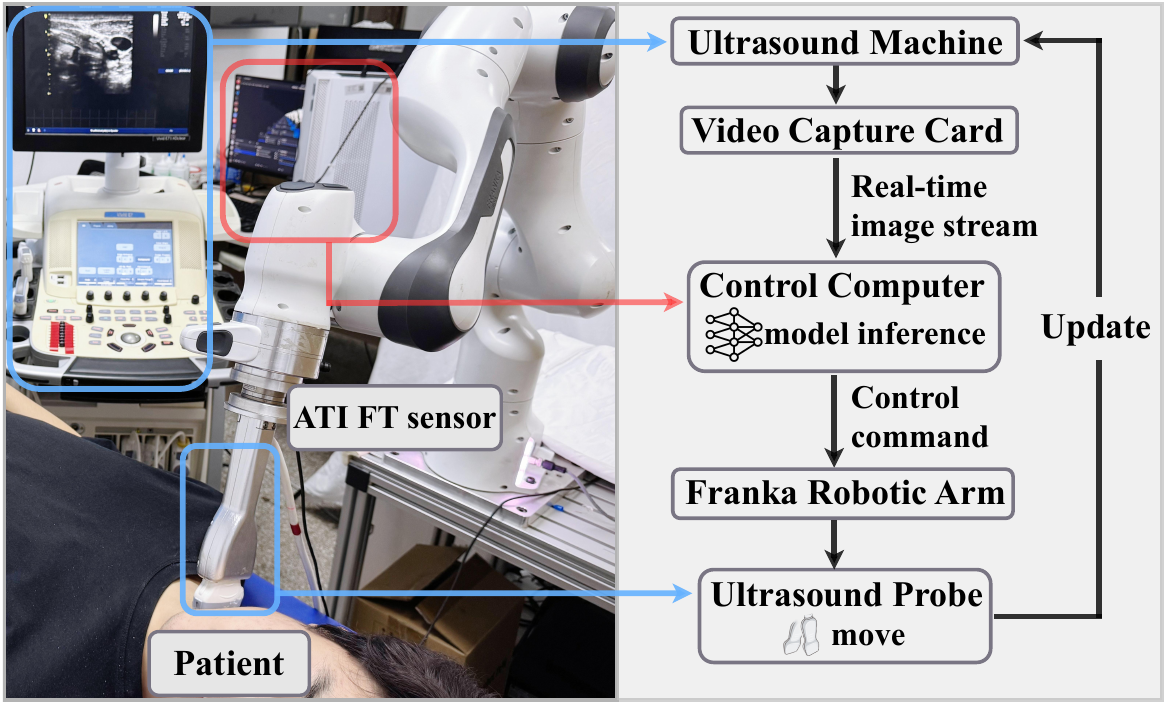}
\caption{Hardware and control configuration of the system.}
\label{robot_setup}
\vspace{-15pt}
\end{figure}

\subsection{Datasets and Implementation Details}

\textbf{Datasets}.
The dataset was collected by sonographers with over 10 years of experience using a GE Vivid E7 ultrasound device equipped with a 9L probe.
During scanning, we synchronously recorded the ultrasound frame and the action taken at each time step.
The data collection details follow the work~\cite{jiang2025towards}.
The corpus contains multiple scans, each scan is a sequence of image--action pairs $\{(\mathbf{I}_t, a_t)\}_{t=1}^T$.
Stage~1 uses data requested from~\cite{jiang2025towards}, which includes 233 scan trajectories with 88{,}712 image--action pairs from 81 subjects (76 for training, 5 for validation).
Stage~2 includes 164 \textbf{newly} collected trajectories with 72{,}279 image--action pairs from 28 subjects (25 for training, 3 for validation).
This study was approved by the institutional ethics committee, and written informed consent was obtained from all participants.

\begin{table*}[!t]\normalsize
\caption{\textbf{Real-world experiment results.}
We report: Success Rate (Final-ICA) – termination on a high-quality ICA longitudinal view; Success Rate (Pass-ICA) – attainment of such a view at any point; Feature Cosine Similarity – similarity to expert scans; Number of Corrections – count of non-routine actions; Subject Comfort Score – participant comfort (0–10).}
\label{tab:real_world}
\centering
\setlength{\tabcolsep}{5pt}
\renewcommand{\arraystretch}{1.15}

\begingroup
\renewcommand{\arraystretch}{1.50} 

\resizebox{2\columnwidth}{!}{
\setlength{\tabcolsep}{0pt} 
\begin{tabular}{L{4.0cm} P{2.6cm} P{2.6cm} P{2.6cm} P{3.0cm} P{3.7cm} P{3.7cm}}
\toprule
\textbf{Method} &
\makecell{\textbf{Success Rate}\\\textbf{(Final-ICA)}} &
\makecell{\textbf{Success Rate}\\\textbf{(Pass-ICA)}} &
\makecell{\textbf{Feature Cosine}\\\textbf{Similarity}} &
\makecell{\textbf{Scan Time (s)}\\\textbf{Mean (min--max)}} &
\makecell{\textbf{Number of Corrections}\\\textbf{Mean (min--max)}} &
\makecell{\textbf{Subject Comfort Score}\\\textbf{Mean (min--max)}} \\
\midrule
Rule-based                        & 20\% (2/10)  & 20\% (2/10)  & 0.4895 & 73.3 (59.6--82.2) & 0 (0--0) & 4.8 (3--6) \\
Rule-based + Explore                & 20\% (2/10)  & 40\% (4/10)  & 0.6124 & 74.5 (54.0--90.9) & 5 (0--10) & 5.2 (5--6) \\
E2E, Single-frame ~\cite{jiang2025towards}   & 35\% (7/20)  & 50\% (10/20)  & 0.6201 & 84.8 (51.9--180) & 5.4 (0--38) & 7.8 (7--9) \\
Hier, Single-frame                & 50\% (10/20) & 70\% (14/20) & 0.6186 & 95.0 (53.8--180) & 9.3 (0--37) & 7.0 (5--9) \\
\textbf{Hier, Sequential (Ours)}  & \textbf{80\% (16/20)} & \textbf{95\% (19/20)} & \textbf{0.7470} & 77.6 (57.4--109.9) & 4.5 (0--16) & 7.6 (6--9) \\
\bottomrule
\end{tabular}
}
\endgroup
\vspace{-5pt}
\end{table*}

\textbf{Model Architecture}.
The UltraHiT is implemented in PyTorch. The Causal Transformer backbones for both the Corrective Gate ($\mathcal{T}_{\mathrm{gate}}$) and Adaptive Corrector ($\mathcal{T}_{\mathrm{corr}}$) utilize a model dimension $D_{\mathrm{model}}=256$ and consist of $L=4$ layers. Each layer contains a Multi-Head Self-Attention mechanism with $h=8$ attention heads and a MLP with an expansion ratio of 4. 
Image features are extracted using a pre-trained DeiT-Tiny backbone, with the resulting $D_{\mathrm{img}}=192$ dimensional features projected to $D_{\mathrm{model}}$ via a linear layer. Action embeddings use $D_{\mathrm{act}}=128$ dimensions before projection. The concatenated image-action features form tokens of dimension $D_{\mathrm{hidden}}=512$. Both models are trained independently using the cross-entropy loss function.

\textbf{Training Strategy}.
We use the Adam optimizer with an initial learning rate of $1\times10^{-4}$ and a cosine learning-rate schedule, weight decay of $1\times10^{-3}$, and a batch size of 256.
All three models are trained with the same recipe for 10 epochs using four NVIDIA A100 GPUs.

\begin{figure}[t!]
\centering
\includegraphics[width=1\columnwidth]{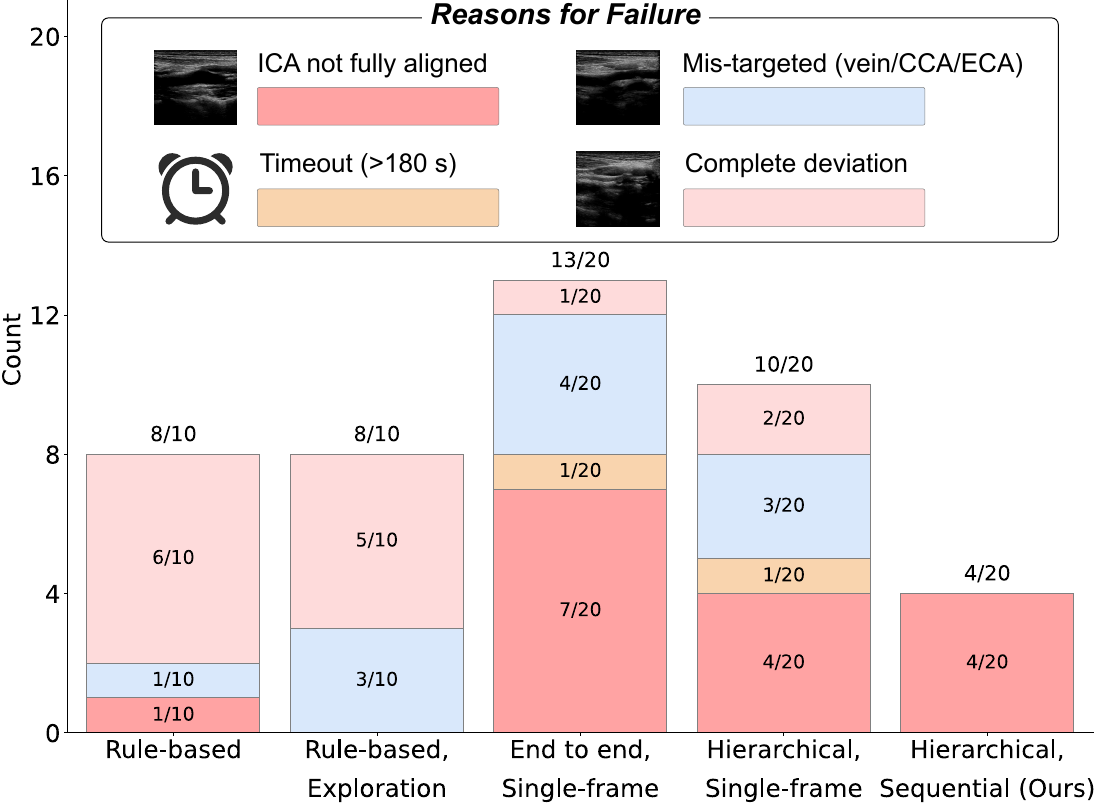}
\caption{Failure reasons for different method.}
\label{fig:fail_reason}
\vspace{-15pt}
\end{figure}

\textbf{Robotic System Configuration}.
We built a robotic ultrasound platform as illustrated in \cref{robot_setup}. The system employs a Franka Emika Panda arm as the manipulator and a GE Vivid E7 ultrasound device fitted with a 9L probe as the imaging source. The probe is rigidly mounted to the end effector with an ATI Mini 40 force/torque sensor for contact force feedback. 
Ultrasound video is captured with an Acasis video capture card and streamed to the control computer for real-time inference.
Before each experiment, an operator applies acoustic gel to the probe surface and roughly places the probe on the participant’s right neck region. The initial positioning requires only approximate placement and does not demand expert precision.
For safety, both hardware and software limits are implemented to prevent unintended motion. The operator can trigger an emergency stop at any time to ensure participant safety.
Each discrete action output by the system occurs at a frequency of approximately 1.5 Hz, while the cartesian impedance controller runs at 1 kHz.

\subsection{Real-World Experiment}
\textbf{Comparison with Baselines}.
Five volunteers who were not included in the training set (3 male and 2 female) participated in the real-world evaluation.
We compared our model against four baselines.
``Rule-based'' uses only the Standardized Path Executor without correction. 
``Rule-based + Explore'' augments the Standardized Path Executor with predefined exploratory motions: after 45 routine steps (a population-average estimate to reach the vicinity of the ICA longitudinal view) in Stage 2, the probe emulates the exploratory fine-tuning maneuvers of sonographers by performing 12 directional searches in anatomically reasonable directions, moving two steps per direction, to locate the ICA longitudinal view.
``End-to-end, Single-frame~\cite{jiang2025towards}'' is a monolithic network predicting motion directly from a single frame.
``Hierarchical, Single-frame'' uses our hierarchical architecture but relies on single-frame input only.
All of these baselines use the same stop model as ours.

For ``Rule-based'' and ``Rule-based + Explore'', each subject performed two trials per method; other methods were tested in four trials per subject. To ensure fair comparison, the probe was manually positioned at a good initial pose: the z-axis of the probe’s coordinate frame was approximately perpendicular to the neck surface, and the common carotid artery was roughly centered in the image. Note that this controlled starting position is only for fairness—Section \ref{robust} later examines robustness under poor initialization.

\begin{figure}[t!]
\centering
\includegraphics[width=1\columnwidth]{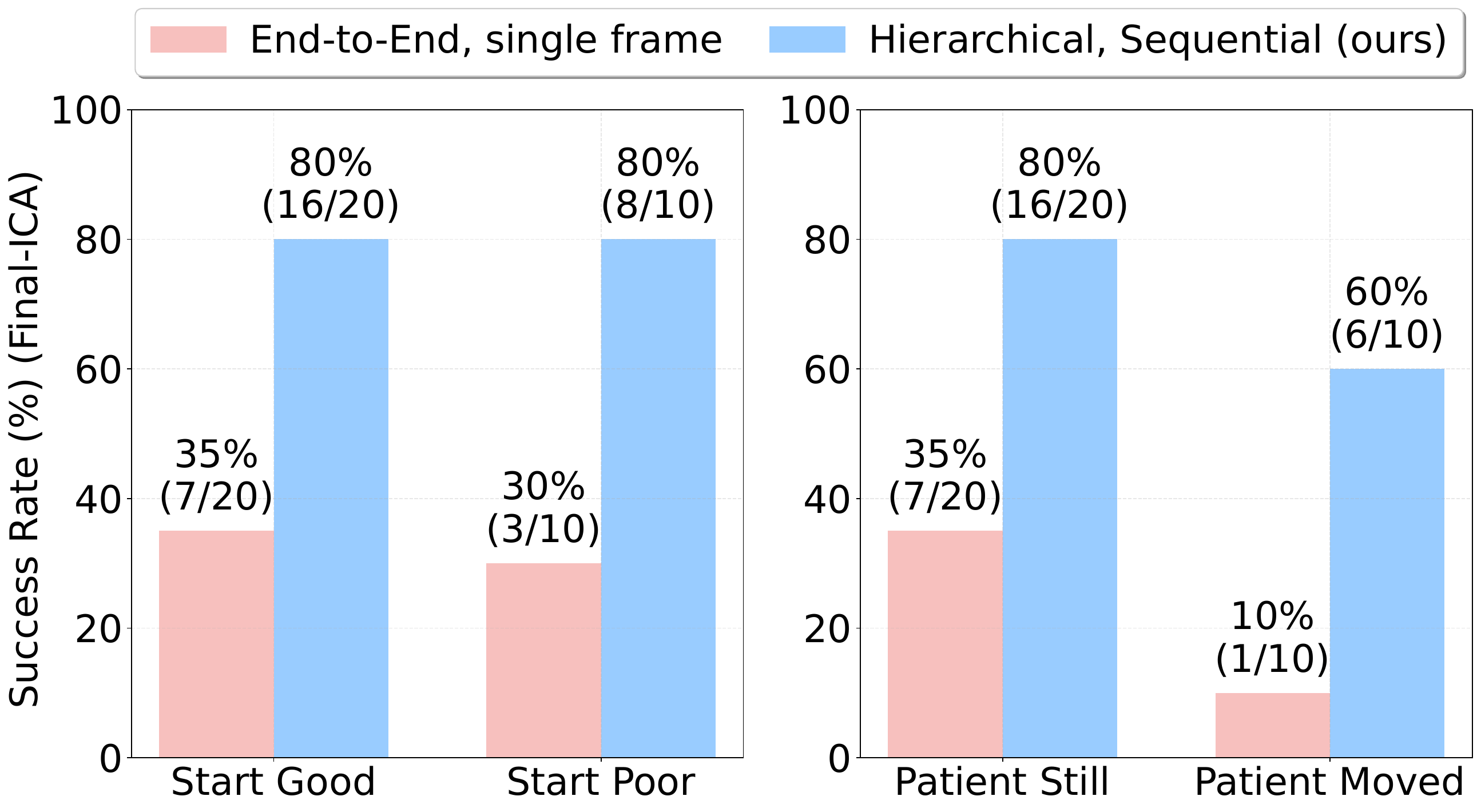}
\caption{Robustness evaluation under atypical conditions.
}
\label{fig:robust}
\vspace{-15pt}
\end{figure}

\begin{figure*}[t!]
\centering
\includegraphics[width=2\columnwidth]{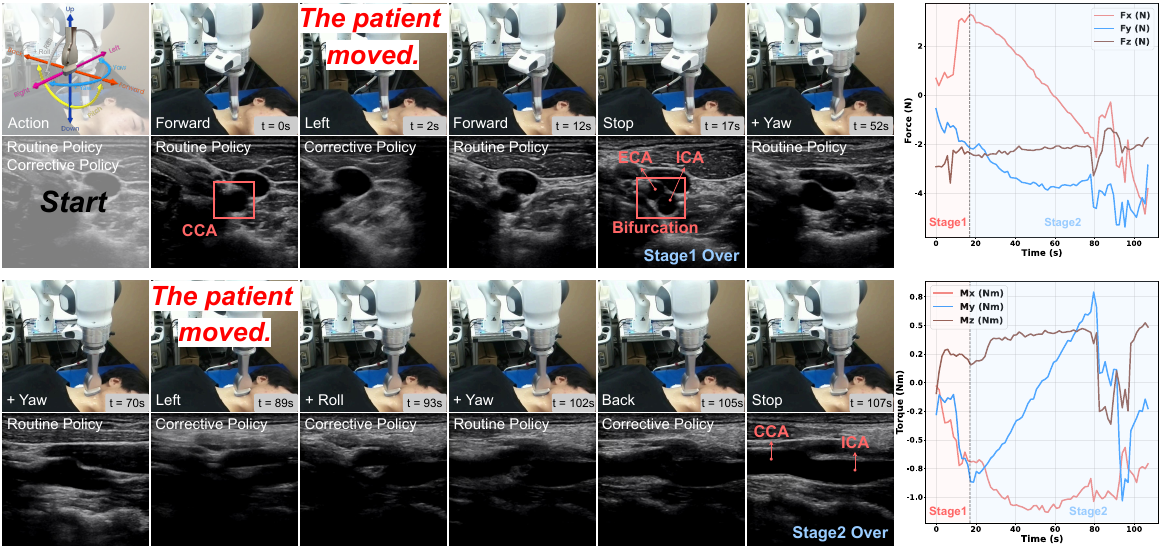}
\caption{A representative scanning sequence where the patient moved twice during scan. Our model accurately applied corrective policy to recover and successfully aligned the ICA longitudinal view.
The right panels plot the real-time force and torque measurements on the ultrasound probe throughout Stage 1 and Stage 2.
}
\label{fig:vis}
\vspace{-10pt}
\end{figure*}

We evaluated six metrics:
(1) Success Rate (Final-ICA) – the probe terminates on a high-quality ICA longitudinal view;
(2) Success Rate (Pass-ICA) – the probe passes through a high-quality ICA longitudinal view at any time;
(3) Feature Cosine Similarity – for each scan, we computed the cosine similarity between all frames’ features and the expert’s final ICA view, taking the sequence maximum as its score. These maxima were averaged across sequences for each method. Higher values indicate closer agreement with expert scans. Features were extracted using the encoder of the pre-trained Stop Model;
(4) Scan Time – duration of each scan, excluding manually aborted trials after complete carotid drift;
(5) Number of Corrections – count of corrective actions per scan, also excluding aborted trials;
(6) Subject Comfort Score – self-reported comfort level (0–10) after each scan.

The results are summarized in \cref{tab:real_world}. Our method achieves the highest success rates (80\% Final-ICA, 95\% Pass-ICA) and feature similarity (0.7470) among all baselines, with competitive scan time and moderate corrections. It also received a high comfort rating (mean = 7.6). Despite requiring deep mandibular insertion, our method's precise, impedance-controlled motions minimized discomfort, demonstrating both effectiveness and safety.

\textbf{Failure Analysis}.
Failure reasons for each method are summarized in \cref{fig:fail_reason}.
During the real-world experiments, ``Rule-based'' and ``Rule-based + Explore'' performed poorly due to difficulty handling anatomical variability of the ICA, with most failures being Complete deviation requiring manual abort.
The ``End-to-end, Single-frame'' model often predicted imprecise motions near the target, resulting in ICA not fully aligned or Mis-targeted vessels (vein/CCA/ECA). It also frequently entered motion loops—repeatedly outputting opposite actions—causing increased scan time and Timeout (\(>180\mathrm{s}\)) failures.
The ``Hierarchical, Single-frame'' model improved corrective prediction but still suffered from action loops and occasional Timeouts.
Our method avoided motion loops through sequence input and achieved higher accuracy. Its four failures were all minor ICA not fully aligned cases.

\begin{table*}[!t]\normalsize
\caption{\textbf{Offline validation results.} We report Accuracy, Precision, Recall, and F1 for each model; mPrec., mRec., and mF1 denote macro-averages over all classes. “Single-frame” indicates the single-frame input paradigm.}
\label{tab:offline_result}
\centering
\setlength{\tabcolsep}{5pt}
\renewcommand{\arraystretch}{1.15}
\resizebox{2\columnwidth}{!}{
\begin{tabular}{P{2.6cm}
                P{1.1cm}P{1.1cm}P{1.1cm}P{1.1cm}
                P{1.1cm}P{1.1cm}P{1.1cm}P{1.1cm}
                | 
                P{1.1cm}P{1.1cm}P{1.1cm}P{1.1cm}}
\toprule
\multirow{2}{*}{\textbf{Method}} 
    & \multicolumn{4}{c}{\textbf{Corrective Gate}} 
    & \multicolumn{4}{c}{\textbf{Adaptive Corrector}} 
    & \multicolumn{4}{|c}{\textbf{Stop Model}} \\
\cmidrule(lr){2-5} \cmidrule(lr){6-9} \cmidrule(lr){10-13}
& Acc. & Prec. & Rec. & F1
& Acc. & mPrec. & mRec. &  mF1
& Acc. & Prec. & Rec. & F1 \\
\midrule
\midrule
\multicolumn{13}{l}{\textit{Stage 1: Locate the bifurcation of ICA and ECA on the transverse section}} \\
Single-frame
& 95.07\% & 69.37\% & 73.15\% & 0.7121
& 87.67\% & 94.07\% & 77.36\% & 0.8299
& \multirow{2}{*}{93.28\%} & \multirow{2}{*}{84.94\%} & \multirow{2}{*}{92.98\%} & \multirow{2}{*}{0.8878} \\
Ours     
& 96.04\% & 78.05\% & 79.15\% & 0.7860
& 90.22\% & 93.30\% & 78.20\% & 0.8260
&  &  &  &  \\
\midrule
\multicolumn{13}{l}{\textit{Stage 2: Switch to the longitudinal section and locate the ICA}} \\
Single-frame
& 84.51\% & 70.62\% & 89.47\% & 0.7893
& 66.72\% & 45.64\% & 63.57\% & 0.5897
& \multirow{2}{*}{93.20\%} & \multirow{2}{*}{84.69\%} & \multirow{2}{*}{90.89\%} & \multirow{2}{*}{0.8768} \\
Ours     
& 88.07\% & 76.52\% & 91.05\% & 0.8316
& 75.09\% & 73.69\% & 66.15\% & 0.6621
&  &  &  &  \\
\bottomrule
\end{tabular}
}
\vspace{-8pt}
\end{table*}

\subsection{Robustness Analysis}
\label{robust}
In practical robotic ultrasound, challenges such as poor initialization, patient movement, coughing, or external disturbances are common and hinder clinical adoption. We evaluated robustness under two conditions: \emph{poor initialization} and \emph{patient moved during scan}.
Tests were conducted on two unseen volunteers, comparing our method with the ``End-to-End, single-frame'' baseline (five trials per condition). \emph{Start Good}: CCA centered with acceptable quality; \emph{Start Poor}: CCA faint and near the edge with low quality. \emph{Patient Still}: neck stationary; \emph{Patient Moved}: deliberate moderate neck shifts once in Stage~1 and once in Stage~2.
As shown in \cref{fig:robust}, initial placement quality had negligible impact on our method, achieving similar Final-ICA success under both start conditions. With patient motion, our method maintained 60\%success versus the baseline’s 10\%, demonstrating superior recovery from disturbances.

\cref{fig:vis} illustrates a case where the subject moves twice; our model activates corrective policy and realigns accurately, confirming robust adaptation suitable for clinical use.

\subsection{Offline Evaluation}

We reports offline validation results for the three models on both stages in Table~\ref{tab:offline_result}. 
Compared with the single-frame baseline, our sequence-based approach achieves consistent improvements in Accuracy, Precision, Recall, and F1 for both the Corrective Gate and the Adaptive Corrector, indicating that temporal context contributes to more reliable decision making. 
The Stop Model uses only single-frame input, as it is sufficient for judgment. 
Experimental results show that it achieves high accuracy and a balanced precision-recall performance in both stages.

\begin{figure}[t!]
\centering
\includegraphics[width=1\columnwidth]{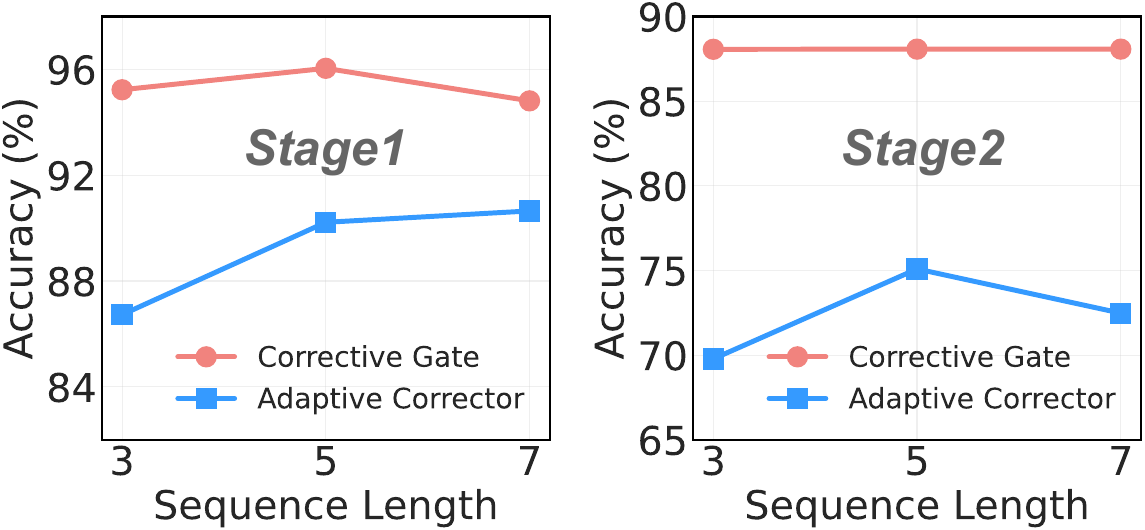}
\caption{Ablation study on sequence length.
}
\label{fig:seq_len}
\vspace{-15pt}
\end{figure}

We also presents an ablation study on sequence length $L\in\{3,5,7\}$ in \cref{fig:seq_len}.
For both stages, longer sequences generally improve accuracy up to a length of 5, after which the benefit saturates or slightly decreases. 
Balancing computational cost and predictive performance, we select a sequence length of 5 for all real-world experiments.

\section{conclusion}

In this work, we present UltraHiT, a hierarchical transformer-based architecture that achieves, for the first time, autonomous scanning of the ICA longitudinal section as required in clinical practice.  By integrating a high-level variation assessment module with two specialized low-level executors—a knowledge-based standard executor and a data-driven adaptive corrector—our method effectively handles significant anatomical variations in the ICA. Experimental results demonstrate that UltraHiT achieves a 95\% success rate on unseen subjects and shows strong robustness in challenging conditions. This work extends the capability of robotic ultrasound to more complex vascular structures and provides a promising framework for handling anatomical variability in medical robotics.

\bibliographystyle{IEEEtran}
\bibliography{reference}

\end{document}